\documentclass{bmvc2k}
\usepackage{booktabs}
\usepackage{wrapfig}

\title{Semantic Segmentation with Active Semi-Supervised Representation Learning}

\addauthor{Aneesh Rangnekar}{aneesh.rangnekar@mail.rit.edu}{1}
\addauthor{Christopher Kanan}{ckanan@cs.rochester.edu}{2}
\addauthor{Matthew Hoffman}{mjhsma@rit.edu}{1}

\addinstitution{
 Rochester Institute of Technology \\
 Rochester, NY, USA
}
\addinstitution{
 University of Rochester \\
 Rochester, NY, USA
}

\runninghead{Rangnekar, Kanan, Hoffman}{S4AL+}

\newcommand{\beginsupplement}{%
        \setcounter{table}{0}
        \renewcommand{\thetable}{S\arabic{table}}%
        
        \setcounter{figure}{0}
        \renewcommand{\thefigure}{S\arabic{figure}}%
    
        \setcounter{equation}{0}
        \renewcommand{\theequation}{S\arabic{equation}}%
        
        \setcounter{section}{0}
        \renewcommand{\thesection}{S\arabic{section}}%
        
        \setcounter{page}{1}
        \renewcommand{\thepage}{S\arabic{page}}%
}

\begin{document}

\maketitle

\begin{abstract}
Obtaining human per-pixel labels for semantic segmentation is incredibly laborious, often making labeled dataset construction prohibitively expensive. Here, we endeavor to overcome this problem with a novel algorithm that combines semi-supervised and active learning, resulting in the ability to train an effective semantic segmentation algorithm with significantly lesser labeled data. To do this, we extend the prior state-of-the-art S4AL algorithm by replacing its mean teacher approach for semi-supervised learning with a self-training approach that improves learning with noisy labels. We further boost the neural network's ability to query useful data by adding a contrastive learning head, which leads to better understanding of the objects in the scene, and hence, better queries for active learning. We evaluate our method on CamVid and CityScapes datasets, the de-facto standards for active learning for semantic segmentation. We achieve more than 95\% of the network's performance on CamVid and CityScapes datasets, utilizing only 12.1\% and 15.1\% of the labeled data, respectively. We also benchmark our method across existing stand-alone semi-supervised learning methods on the CityScapes dataset and achieve superior performance without any bells or whistles.

\end{abstract}

\section{Introduction}


Getting labels for supervised learning problems is challenging, especially for semantic segmentation where these labels are needed on a per-pixel level. The most widely used methods for reducing the need for labels are semi-supervised learning (SSL) and pool-based active learning (AL). Semi-supervised learning proposes strategies to use the unlabeled dataset alongside labeled samples, typically by maintaining an exponential moving average of the network to predict pseudo labels \cite{hu2021semi,wang2022semi,liu2021perturbed}, and pool-based active learning queries the most informative samples within the unlabeled data pool, in terms of pixels, regions, or entire images, to add to the labeling pool via a predetermined scoring mechanism \cite{equal2020,Casanova2020Reinforced,Shin_2021_ICCV,rangnekar2022semantic}. Recently, both EquAL \cite{equal2020} and S4AL \cite{rangnekar2022semantic} have pioneered combining active and semi-supervised learning for semantic segmentation, resulting in significant gains compared to using either in isolation. However, these methods are complicated - the former relies on enforcing consistency in the predictions without accounting for their correctness, and the latter relies on complex data augmentation schemes for convergence. Here we propose S4AL+, which greatly simplifies the S4AL method by replacing its mean-teacher framework with self-training, thereby eliminating the use of complex data augmentations \cite{xie2020self,yang2021st++}.

\begin{figure}
    \centering
    \includegraphics[width=0.95\linewidth]{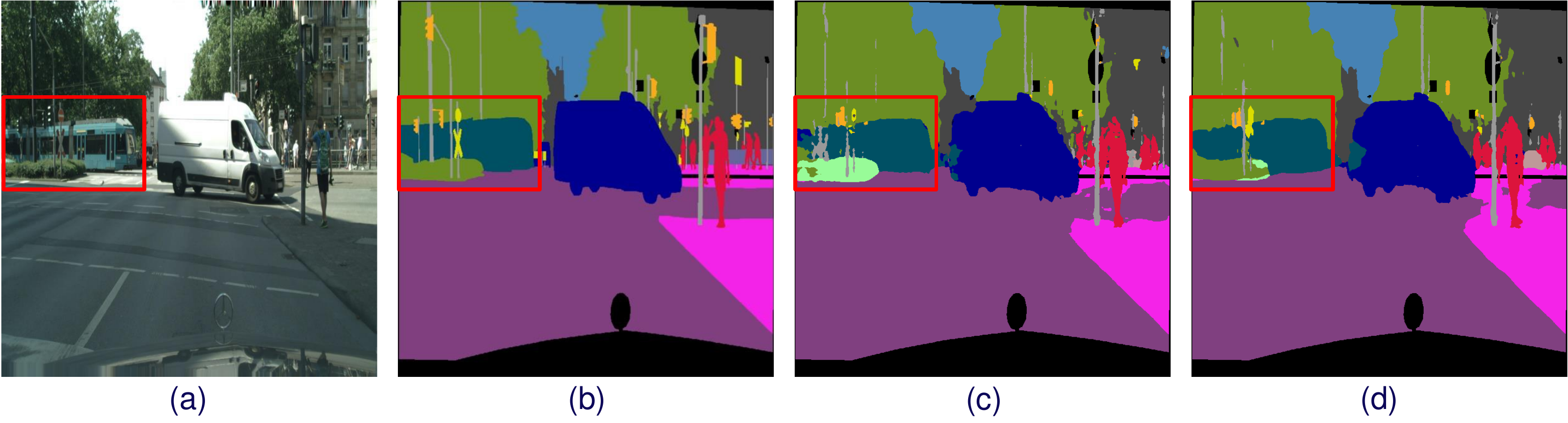}
    \caption{An example from the CityScapes dataset for Active learning: (a) Shows the image, (b) The ground truth labels, (c) The predictions from a network trained only with cross-entropy loss, and (d) The predictions from a network trained with contrastive representation learning. The red box highlights the improved predictions in for the train and the nearby vegetation over standard training with cross-entropy loss.}
    \label{fig:datasets_al_selling}
\end{figure}

Most methods for active learning rely on cross-entropy loss for training their networks \cite{xie2020deal,equal2020,rangnekar2022semantic}. While cross-entropy is a widely used loss function in semantic segmentation, it operates on a individual pixels of the semantic map and fails to take into consideration possible cues of similarity and differences in areas throughout the entire image or set of images to strengthen its learning further (Fig. \ref{fig:datasets_al_selling}). The computer vision community has developed several algorithms to address this problem, notably Conditional Random Fields (CRF) \cite{Chen2018DeepLabSI}, Affinity Nets \cite{liu2017learning}, Region Mutual Information Loss (RMI) \cite{zhao2019region} and Contrastive Learning \cite{zhang2021looking,liu2021bootstrapping}. We hypothesize that boosting cross-entropy performance with any of these methods can improve the quality of pseudo labels, thereby improving the quality of queried labeling instances. 

In this work, we show how active learning for semantic segmentation can be improved with a straightforward technique: at each active learning cycle, we produce pseudo labels for the dataset with the self-training framework \cite{xie2020self} and leverage contrastive representation learning to improve the boundaries between different classes \cite{liu2021bootstrapping}. Our combination of self-training with contrastive representation learning in S4AL+ enables querying superior samples for active learning, which results in more efficient learning. While S4AL+ was designed for active learning, we also demonstrate state-of-the-art performance on semi-supervised learning. 

\section{Background}

\textbf{Pool-based Active Learning for Semantic Segmentation} is a technique for ranking unlabeled data points on their importance based on machine learning methods including, but not limited to, consistency \cite{li2020uncertainty,siddiqui2020viewal,equal2020,haussmann2020scalable}, diversity \cite{sinha2019variational,ebrahimi2020minimax} and feature level learning \cite{Casanova2020Reinforced,xie2020deal,Shin_2021_ICCV,rangnekar2022semantic}. These frameworks can be further classified based on their approach for querying the data (image \cite{haussmann2020scalable,xie2020deal,sinha2019variational,ebrahimi2020minimax}, region \cite{li2020uncertainty,siddiqui2020viewal,equal2020,Casanova2020Reinforced,rangnekar2022semantic} or pixel \cite{Shin_2021_ICCV}). Notably, EquAL \cite{equal2020}, Minimax \cite{ebrahimi2020minimax}, and S4AL \cite{rangnekar2022semantic} proposed including the unlabeled dataset as part of the training protocol to achieve better image-level (Minimax) or region-level (EquAL, S4AL) results. Unlike these approaches, we show that adding a simple contrastive embedding head and self-training to boost the networks' quality of pseudo labels is sufficient to attain state-of-the-art results.

\textbf{Semi-Supervised Learning for Semantic Segmentation} commonly learns representations using a teacher-student framework \cite{mittal2019semi,french2020semisupervised,zhu2020improving,xie2020self,Olsson_2021_WACV,hu2021semi,Guan_2022_CVPR}, and recently, combining teacher-student with contrastive embeddings \cite{liu2021bootstrapping,zhou2021c3,alonso2021semi,wang2022semi}. Most teacher-student frameworks use the mean teacher framework as their foundation \cite{antti2017meanteacher}, wherein a slowly updated version (teacher) of the continuously updated model (student) predicts reliable pseudo labels on the unlabeled data points for joint training. The exceptions to this line of approach are ST++ \cite{yang2021st++} and USRN \cite{Guan_2022_CVPR} which use self-training based learning \cite{xie2020self}. These algorithms complement our work, which differs in two aspects of learning. Unlike ST++, we use pixel-level (local) metrics instead of image-level (global) metrics for determining the quality of pseudo labels, and unlike USRN, we use contrastive representation learning to improve the quality of pseudo labels instead of clustering to improve the data distribution.

\textbf{Supervised Contrastive Learning} stems from adapting the one-to-many contrastive loss to the many-to-many pipeline for mapping similarities and differences between data points in the representation space \cite{khosla2020supervised}. Multiple frameworks aim to boost the pixel-wise predictions in semantic segmentation by incorporating contrastive learning with memory banks (prototypes) for supervised learning \cite{zhang2021looking,Wang_2021_ICCV,zhou2022rethinking} and semi-supervised learning \cite{zhou2021c3,alonso2021semi,wang2022semi}. Our work employs a much lighter technique, ReCo \cite{liu2021bootstrapping}, which continuously updates the contrastive embeddings per data iteration and achieves a comparative performance to its memory banks counterparts.

\section{Our Active Learning Framework \& Algorithm: S4AL+}

Active learning methods work with an initial labeled pool ($D_L$, $x_l$) and a relatively larger unlabeled pool ($D_U$, $x_u$), intending to query instances from $D_U$ for labeling. This task is performed at every active learning cycle (${AL}_T$), whose stopping criteria (${AL}_C$) is either preset or determined as the number of cycles required to achieve comparable performance to supervised learning ($D_L$, $D_U \longrightarrow D_L$).

We build S4AL+ on the foundations of S4AL \cite{rangnekar2022semantic}, self-training \cite{xie2020self} and contrastive representation learning \cite{liu2021bootstrapping}. Following S4AL, we query the instances $x_u$ at every ${AL}_T$ at region level using the average entropy of the network prediction's in the region. At the end of every active learning cycle, we query for labeled data from $D_U$ by ranking the regions according to network's entropy and we account for pseudo labels by considering every pixel whose prediction confidence is above a fixed threshold. In this manner, the network has richer labels within the unlabeled pool at the beginning of the next active learning cycle and does not need to iteratively predict pseudo labels every data iteration \cite{antti2017meanteacher,rangnekar2022semantic}.

\begin{figure}[t]
    \centering
    \includegraphics[width=0.8\linewidth]{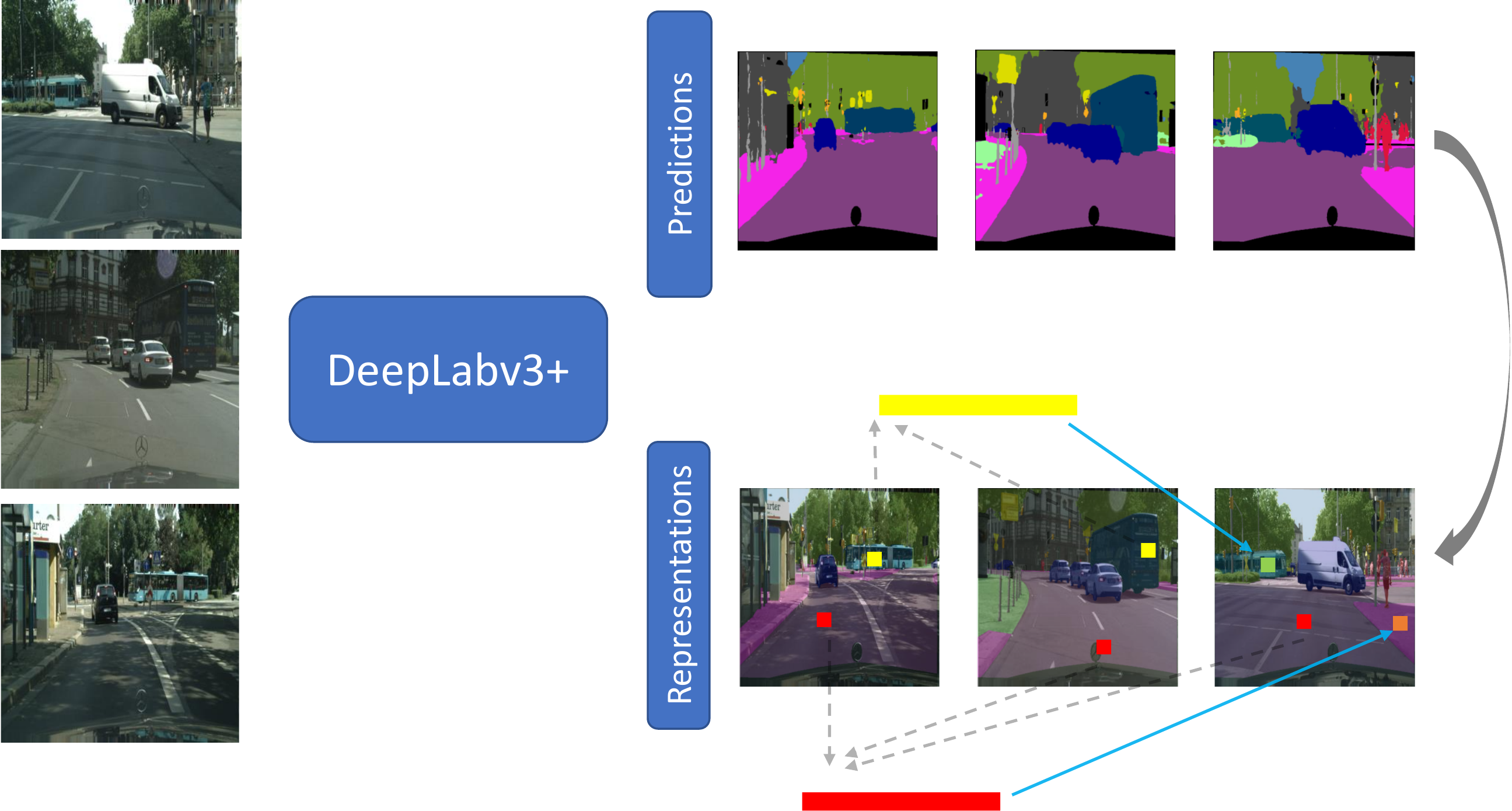}
    \caption{Schematic showing how our modified DeepLabv3+ framework employs contrastive representation learning to improve the performance of pseudo labels per learning cycle, which improves the performance of our active learning framework. Note how road keys group together (red) and ensure distance from sidewalk (orange), and similarly, bus keys (yellow) separate from train (green). The confidence in the prediction map highlights the keys for smart selection during contrastive learning (black arrow).}
    \label{fig:mainalgo}
\end{figure}

To ensure high quality within the queried regions, we investigate allowing the network to `peep' into the data within the unlabeled pool by using semi-supervised learning. We achieve this with two simple adjustments: 1) We train the network with the self-training framework \cite{antti2017meanteacher,xie2020self}, and 2) We add a contrastive embedding head to strengthen class-wise representations \cite{liu2021bootstrapping} (Fig. \ref{fig:mainalgo}). We briefly discuss the motivations and functionality of both our modifications, before discussing our findings.

\subsection{Self-training}

\textbf{Motivation.} Mean-teacher based approaches for semi-supervised learning, wherein an exponentially moving average of the student network is used to predict pseudo labels on $D_U$, requires complex data augmentations (variants of CutMix \cite{yun2019cutmix} or ClassMix \cite{Olsson_2021_WACV}) and regularizations to function successfully on semantic segmentation. Multiple research studies in the areas of semi-supervised learning and active learning (\cite{hu2021semi,wang2022semi,rangnekar2022semantic}) use modified data augmentation pipelines to ensure robust pseudo labels. On the contrary, self-training uses a relatively simple data augmentation scheme, while managing to achieve comparable performance.

The self-training pipeline for semi-supervised semantic segmentation can be summarized in the following steps:
\begin{enumerate}
    \item\label{item:st_step1} Train the network $N_T$ on ${D_L}^{wa}$ with $CE$,
    \item\label{item:st_step2} Obtain pseudo labels for $D_U$ from $N_T$  with a determined threshold, 
    \item\label{item:st_step3} Jointly train ${D_L}^{wa}$ and ${D_U}^{sa}$ on $N_S$ with $CE$,
    \item\label{item:st_step4} $N_S \longrightarrow N_T$,
    \item\label{item:st_step5} Repeat 2 to 4 till convergence,
\end{enumerate}
where $N_S$, $N_T$, $wa$, $sa$, $CE$ indicate student network, teacher network, weak data augmentation, strong data augmentation, and cross-entropy loss respectively. Self-training also relies on injecting noise into the network for augmentations, which we achieve with Dropout for MobileNetV2 and Stochastic Depth for the ResNet variants in our framework \cite{srivastava14a,huang2016deep,Sandler_2018_CVPR,He_2016_CVPR}. This pipeline can be retrofitted for active learning with two minor adjustments. First, we modify Step \ref{item:st_step5} to repeat itself every ${AL}_T$, until we reach the stopping criteria cycle ${AL}_C$. We also query $x_u$ within $D_U$ for labeling during every iteration of Step \ref{item:st_step2}. As a result, we obtain the relatively easier set of labels in $D_U$ via pseudo labeling and the harder set of labels in $D_U$ via manual annotations at each active learning cycle (${AL}_T$). This results in a larger portion of information available for the network during the next training phase of the active learning cycle (${AL}_T$).

\subsection{Contrastive Representation Learning}

\textbf{Motivation.} We face two challenges when directly applying the framework for semantic segmentation: 1) for classification, the entire feature representation for the network corresponds to a single image category, and 2) cross-entropy loss works on a per-pixel basis and ignores the possibility of learning from other pixels in the image (and the batch) to improve its understanding. Hence, we explore the field of supervised contrastive learning and augment our network with a contrastive embedding head for learning meaningful representations (\ref{fig:mainalgo}). Specifically, we adapt the ReCo loss \cite{liu2021bootstrapping} into our training pipeline due to its simplicity and lower memory overhead.

We modify our network for semantic segmentation by adding another decoder head that accounts for learning representations and is trained via the ReCo loss (Fig. \ref{fig:mainalgo}). Assuming that $r$ indicates the representations obtained in parallel to the network's class-wise predictions, and $C$ is the total number of classes present in the dataset during the mini-batch under training,  Eqn. \ref{loss:reco} summarizes the fundamentals of the ReCo loss \cite{liu2021bootstrapping} as follows: 
\begin{equation}
  L_{\tt ReCo} = \sum_{c\in \mathcal{C}} \sum_{r_q \sim \mathcal{R}^c_q} -\log \frac{\exp(r_q \cdot r_k^{c, +} / \tau)}{\exp(r_q \cdot r_k^{c, +}/ \tau) + \sum_{r_k^{-}\sim \mathcal{R}^c_k} \exp(r_q \cdot r_k^{-}/ \tau)},
  \label{loss:reco}
\end{equation}
where 
\begin{itemize}
    \item $\mathcal{R}_q^c$ represents the positive set containing all representations whose ground truth labels belong to class $c$,
    \item $r_k^{c, +}$ represents the positive anchor, which is the mean of $r_q$ per class $c$,
    \item $r_k^{-}$ are all representations within ${R}^c_k$, which is the negative key set containing all representations whose ground truth label is not class $c$, and 
    \item $\tau$ is the scalar temperature control parameter.
\end{itemize}
Eqn. \ref{loss:reco} accounts for all positive and negative keys present within a mini-batch whose size can grow exponentially based on the number of images and their resolutions. ReCO alleviates this problem by constructing a pair-wise similarity graph among the mean representations of every class ($r_k^{c, +}$, $c$ $\in$ $C$) and actively sampling for a meaningful set of negative keys $r_k^{-}$ per $\mathcal{R}_q^c$, whose quantity is a tunable hyper-parameter (${RC}_{K}$). This ensures that semantically different classes, for example, Bus and Vegetation, are rarely sampled as a pair for learning, while semantically similar classes like Bus and Train are seen more often.

Similarly, $r_q$ are chosen based on the threshold ${RC}_{\delta_s}$ for every corresponding prediction confidence from the network and a tunable hyper-parameter (${RC}_{Q}$) which determines the quantity, similar to ${RC}_{K}$. ${RC}_{\delta_s}$ ensures that positive keys belong to areas within the prediction map which can benefit from representation learning. In this manner, ReCo ensures that only the most informative sets (both positive and negative) are trained while maintaining a minimal computation overhead. We replace the CE loss function in Steps \ref{item:st_step1} and \ref{item:st_step3} with a combination of CE and ReCo loss to further boost the network's learning potential.

\subsection{Summary: S4AL+ vs. Other Methods}

Our approach differs from S4AL, and other approaches in the semi-supervised and active learning literature in three ways: 1) We do not use the mean teacher pipeline to predict pseudo labels at every mini-batch iteration, in this manner, we save on the training time and also potential bias from the labels imbalance being carried forward every step, 2) We do not rely on heavy augmentations like CutMix \cite{yun2019cutmix} or ClassMix \cite{Olsson_2021_WACV}, and 3) We use contrastive learning with self-training to utilize the unlabeled pool to its maximum potential (Fig. \ref{fig:datasets_al_selling}). 

\section{Experiments and Results}

Before presenting our results, we first introduce the metrics and datasets used in experiments. 
\subsection{Datasets, metrics and training protocol}

\paragraph{Datasets.}
We conduct experiments on the CamVid, and CityScapes datasets \cite{brostow2009semantic,cordts2016cityscapes}. CamVid has a default resolution of 720 $\times$ 960, with 11 classes of interest, and CityScapes has a default resolution of 1024 $\times$ 2048, with 11 classes of interest. For active learning, we downsample CamVid and CityScapes to 360 $\times$ 480 and 688 $\times$ 688, respectively, for training and evaluation, following \cite{xie2020deal,rangnekar2022semantic}. For semi-supervised learning, we crop images within the CityScapes dataset at 768 $\times$ 768 for training, and maintain the original resolution for evaluation, following \cite{hu2021semi,wang2022semi}.

\paragraph{Metrics.}
We evaluate our algorithm's performance for active learning by measuring the proportion of additional labeled data required to reach more than 95\% performance in terms of mean IoU (mIoU) on fully supervised learning. We evaluate our algorithm's performance for semi-supervised learning by measuring the mIoU achieved using a specific \% of fully supervised data as the labeled data.

\paragraph{Network Architectures.}
We use MobileNetV2 \cite{Sandler_2018_CVPR} with a modified stride of 16 as the backbone for all experiments on active learning (following \cite{xie2020deal,rangnekar2022semantic}), except in comparison to EquAL \cite{equal2020}, wherein we use ResNet-50 with a modified stride of 8. We also use ResNet-50 \cite{He_2016_CVPR} and ResNet-101-DeepStem \cite{he2019bag}, with modified stride of 8, for experiments on semi-supervised learning as per the protocol in the comparative state-of-the-art. We use the DeepLabv3+ framework for semantic segmentation throughout all our experiments \cite{Chen_2018_ECCV}.

\paragraph{Network Optimization.}
We begin all our experiments with an initial learning rate of 1$\times10^{-2}$ and use the ``poly''  learning schedule to gradually decrease the learning rate, similar to \cite{Chen_2018_ECCV,xie2020deal,rangnekar2022semantic}. We use SGD for optimization and a weight decay of 0.0001 in all our experiments. We train the networks for 100 epochs on the CamVid and CityScapes datasets for active learning, and 240 epochs on the CityScapes dataset for semi-supervised learning. For all data augmentation schemes, we use random application of image resizing and horizontal flipping as the weak augmentations set ($wa$), and Gaussian blur, Color Jitter, and CutOut \cite{devries2017improved} as the strong augmentations set ($sa$). We do not create our own splits, but instead use the data splits provided in individual benchmarks to minimize disparity in our findings. Tables \ref{tab:trainingschedule_al} and \ref{tab:trainingschedule_ssl} contain the details of our training procedures on active learning and semi-supervised learning respectively.

\begin{figure}[t]
    \centering
    \includegraphics[width=0.95\linewidth]{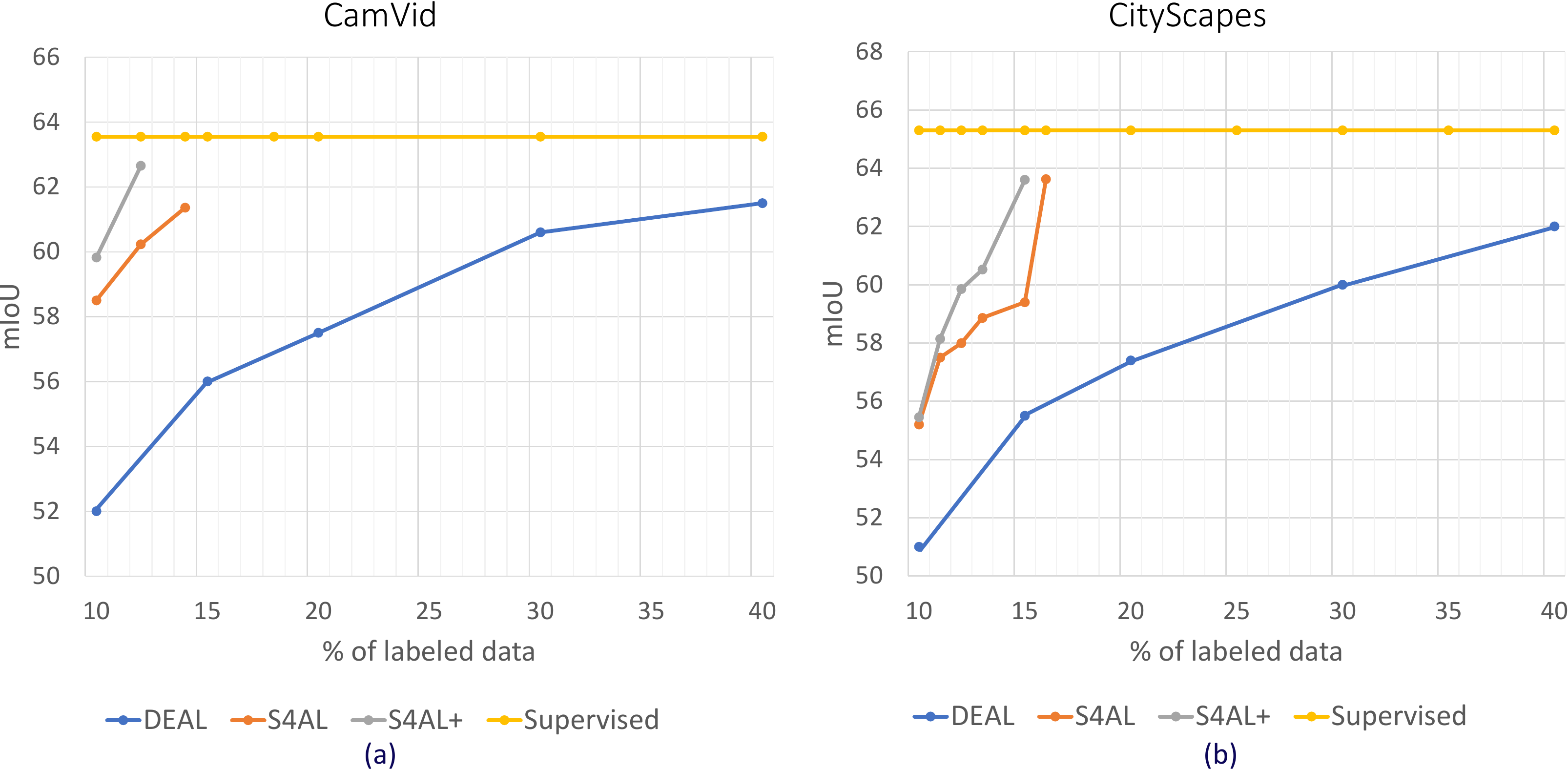}
    \caption{We demonstrate that our framework, S4AL+, works the best for active learning on the CamVid (a) and CityScapes (b) datasets. Both graphs show the mIoU improvements relative to the amount of labeled data utilized in comparison with the previous state-of-the-art frameworks, DEAL \cite{xie2020deal} and S4AL \cite{rangnekar2022semantic}, and the supervised learning performance.}
    \label{fig:results_graphs_activelearning}
\end{figure}

\begin{figure}[t]
    \centering
    \includegraphics[width=0.8\linewidth]{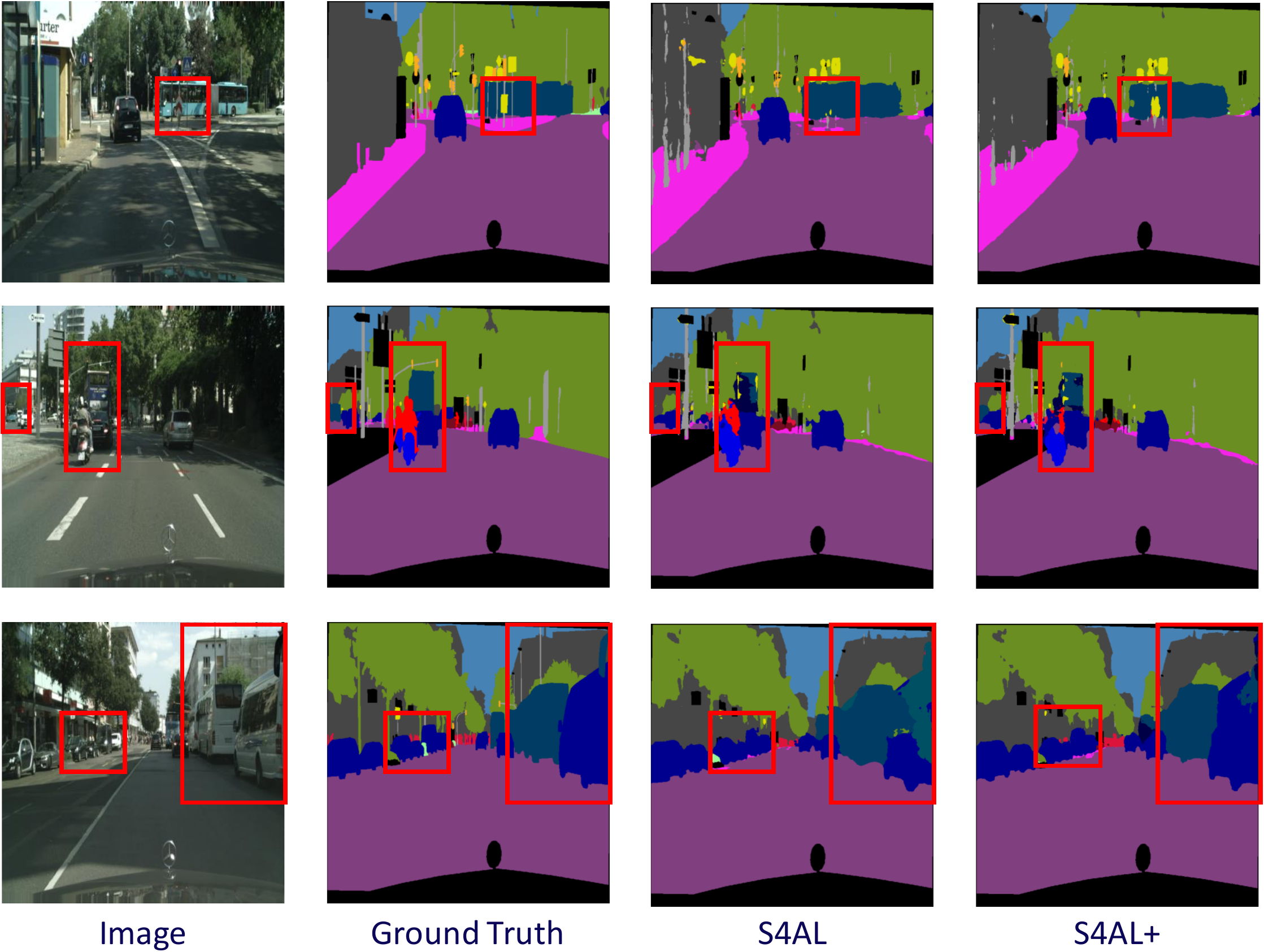}
    \caption{Comparison of our framework to the previous state-of-the-art S4AL. The areas of interest are highlighted and refer to regions wherein our approach supersedes S4AL in terms of correct pixel prediction.}
    \label{fig:results_outputs_activelearning}
\end{figure}

\subsection{Results: Active Learning}

\begin{table}[]
\centering
\def\arraystretch{1.2}
\caption{Summary of all Active learning experiments on CamVid and CityScapes datasets. LD and ULD indicate the percentage of labeled data and unlabeled data respectively. We observe that using 1:2 labeled to unlabeled image ratio, and training for longer epochs, on the final active learning cycle, result in superior gains with respect to mIoU on both datasets.}
\label{tab:trainingschedule_al}
\resizebox{\columnwidth}{!}{%
\begin{tabular}{llllllll}
Dataset & Experiment & Epochs & \begin{tabular}[c]{@{}l@{}}Batch\\ Size\\ (Labeled)\end{tabular} & \begin{tabular}[c]{@{}l@{}}Batch\\ Size\\ (Unlabeled)\end{tabular} & \begin{tabular}[c]{@{}l@{}}Iterations\\ Per Epoch\end{tabular} & \begin{tabular}[c]{@{}l@{}}Contrastive\\ Representation\\ Learning?\end{tabular} & mIoU \\ \midrule
CamVid & 10\% LD & 100 & 4 & - & 50 &  & 58.42 \\
 & 10\% LD & 100 & 4 & - & 50 & Yes & 60.11 \\
 & 10\% LD, ULD & 100 & 2 & 2 & 100 & Yes & 60.50 \\
 & 10\% LD, ULD & 100 & 2 & 4 & 100 & Yes & 60.92 \\
 & 10\% LD, 2.1\% LD, ULD & 100 & 2 & 4 & 100 & Yes & 61.80 \\
 & 10\% LD, 2.1\% LD, ULD & 150 & 2 & 4 & 100 & Yes & 62.65 \\ \midrule
CityScapes & 10\% LD & 100 & 4 & - & 100 &  & 54.84 \\
 & 10\% LD & 100 & 4 & - & 100 & Yes & 55.45 \\
 & 10\% LD, ULD & 100 & 2 & 2 & 300 & Yes & 56.31 \\
 & 10\% LD, ULD & 100 & 2 & 4 & 300 & Yes & 57.12 \\
 & 10\% LD, 1.3\% LD, ULD & 100 & 2 & 4 & 300 & Yes & 58.14 \\
 & 10\% LD, 2.6\% LD, ULD & 100 & 2 & 4 & 300 & Yes & 59.85 \\
 & 10\% LD, 3.8\% LD, ULD & 100 & 2 & 4 & 300 & Yes & 60.52 \\
 & 10\% LD, 5.1\% LD, ULD & 100 & 2 & 4 & 300 & Yes & 61.47 \\
 & 10\% LD, 5.1\% LD, ULD & 200 & 2 & 4 & 300 & Yes & 63.60 \\ \bottomrule
\end{tabular}
}
\end{table}

We conduct experiments for the full active learning system with the CamVid and CityScapes datasets with MobileNetV2 encoder and DeepLabv3+ framework. In both cases, we begin with 10\% of the data as our labeled dataset and gradually query for additional labeled data per active learning cycle (\cite{xie2020deal,rangnekar2022semantic}). We query four regions per image of 30 $\times$ 30 and 43 $\times$ 43 for each active learning cycle on CamVid and CityScapes, respectively, following \cite{rangnekar2022semantic}. We also assign pseudo labels to the pixels that meet our prediction confidence threshold of $0.7$ in every image in the unlabeled data pool. 

Fig. \ref{fig:results_graphs_activelearning} shows our results on the CamVid and CityScapes dataset. Our approach achieves over 95\% of the full dataset performance with only 12.1\% of the labeled pixel data for the CamVid dataset, and 15\% labeled pixel data on the CityScapes dataset. We observe a significantly larger boost in the performance during the final active learning cycles, as we allow the network to train for a longer time, motivated by  \cite{beyer2022knowledge,rangnekar2022semantic}. Fig. \ref{fig:results_outputs_activelearning} shows the difference in outputs from S4AL and S4AL+, highlighting the advantage of using contrastive learning to learn better object features. We refer the reader to Tables~\ref{tab:activelearning_classwise_camvid} and  \ref{tab:activelearning_classwise_cityscapes} in the supplemental for a discussion on the class-wise performance on both datasets.

We further compare our system to EquAL \cite{equal2020}, another region-based selection method, that uses a ResNet-50 encoder with DeepLab-v3+ \cite{He_2016_CVPR,Chen_2018_ECCV}. Starting with 8\% labeled data and a budget of 12\% labeled data on the CamVid dataset, our approach achieved an mIOU of 66.4 on CamVid, compared to 63.4 from EquAL and 65.3 from S4AL. When starting with 3.5\% data on the CityScapes dataset, our approach achieves a mIoU of 67.5 with only 10\% labeled data, compared to EquAL's 67.4 with 12\% data and S4AL's 66.7 with 10\% data, thus demonstrating superior performance of our method on multiple training processes.

\subsection{Results: Semi-supervised Semantic Segmentation}

\begin{figure}[t]
    \centering
    \includegraphics[width=0.9\linewidth]{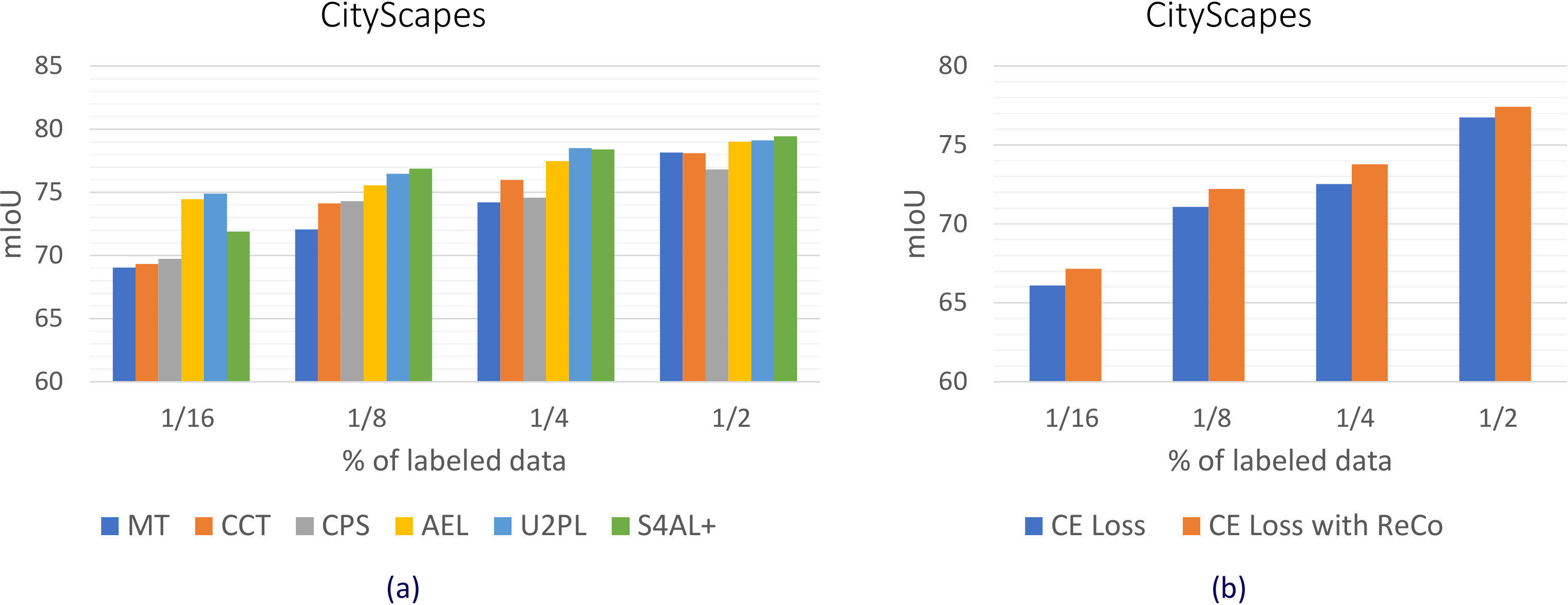}
    \caption{(a) Comparison of our framework in terms of mIoU to the previous state-of-the-art approaches for semi-supervised learning on CityScapes dataset. (b) Comparison of improvements gained by using contrastive learning (ReCo \cite{liu2021bootstrapping}) for semi-supervised learning on CityScapes dataset.}
    \label{fig:results_graph_semisupervisedlearning}
\end{figure}

We compare our approach without the active learning pipeline to semi-supervised learning methods. Specifically, we do not query for any additional data from the unlabeled pool of images and only use ground truth for those images that fall in the training set. We train the network on the labeled training data, assign the pseudo labels based on the prediction confidence, and retrain the network jointly with labeled and (pseudo-labeled) unlabeled data twice, before reporting our results.

For semi-supervised Learning, we vary the percentage (\%) of labeled images within training data. Fig. \ref{fig:results_graph_semisupervisedlearning}(a) shows the performance of our framework in terms of previous state-of-the-art approaches. We compare S4AL+ with Mean Teacher \cite{antti2017meanteacher}, Cross Consistency Training (CCT) \cite{Ouali_2020_CVPR}, Cross Pseudo Supervision (CPS) \cite{Chen_2021_CVPR}, Adaptive Equalized Learning (AEL) \cite{hu2021semi} and Unreliable Pseudo Labels (U$^2$PL) \cite{wang2022semi} which use the ResNet-101-DeepStem encoder. We observe a slight drop in the performance on the $1/16$ labeled data scenario, but perform at-par or better than other approaches in all other scenarios. Our method is also superior in terms of efficiency as 1) it does not use multiple sets of encoders and decoders (like CCT and CPS), 2) does not complex memory bank mechanisms to actively sample for long tail distribution classes (MT, AEL and U$^2$PL), and 3) does not actively obtain pseudo labels at every data iteration (MT, AEL and U$^2$PL). We refer the readers to Table~\ref{tab:semisupervisedlearning_cityscapes} in the supplemental for in-depth results on the CityScapes dataset.

\begin{figure}[t]
    \centering
    \includegraphics[width=0.95\linewidth]{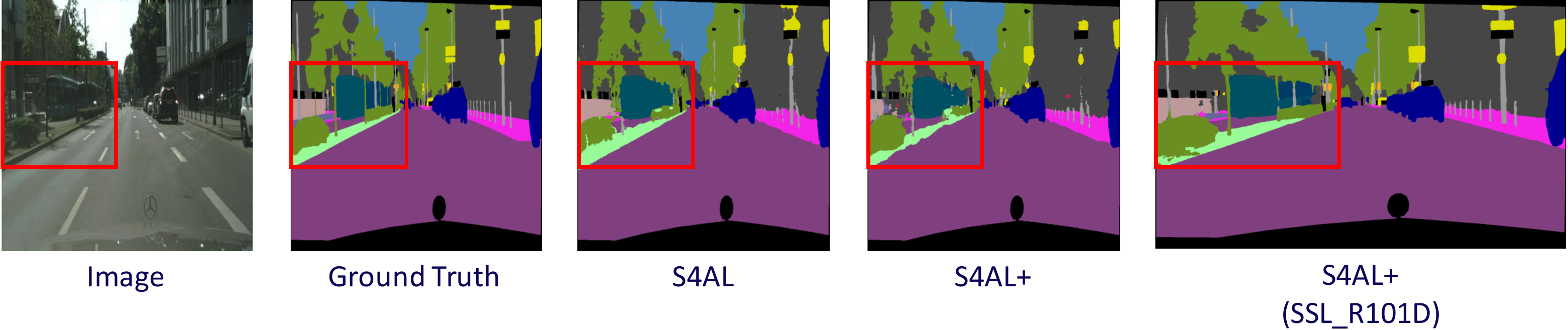}
    \caption{Knowledge distillation? S4AL+ still makes incorrect predictions under cases of extra-similar object appearances. However, we show that this is a limitation of the encoder, MobileNetV2, by comparing its outputs post active learning to the stand-alone output of the same approach trained under the semi-supervised learning framework (with a similar amount of initial labeled data) and the significantly more powerful ResNet-101-DeepStem encoder.}
    \label{fig:results_al_limitations}
\end{figure}

\section{Discussion}

Our results enable us to answer two questions:
\begin{enumerate}
    \item \textit{Does self-training help the case of active learning?} Yes. As a stand-alone adjustment, self-training boosts the initial mIoU by 1.8 points on the CamVid dataset and 1.2 points on the CityScapes dataset when using the MobileNetV2 encoder.
    \item \textit{Does representation learning help the case of active learning?} Yes again. We observe a boost of 2 points in the initial mIoU for the CamVid dataset and 0.8 points on the CityScapes dataset with the MobileNetV2 encoder. We also verify this for semi-supervised learning in Fig. \ref{fig:results_graph_semisupervisedlearning}(b) and Table \ref{tab:semisupervisedlearning_cityscapes}, wherein we use variants of a stronger ResNet encoder.
\end{enumerate}

\section{Conclusion}

We propose a solution for improving active learning for semantic segmentation motivated by the desire to eliminate dependency on data augmentation schemes that involve randomness to ensure robust pseudo labels. We achieve this by formulating active learning as a step-wise semi-supervised learning problem, using self-training compared to the popular teacher-student-based framework. We enrich this framework with a representation learning head that ensures the network can maximize its learning potential. While both our components seem reasonably straightforward, we emphasize that our goal was not to create something entirely new but to develop a solution from the existing literature. S4AL+ is a strong, but elegant, method for active and semi-supervised learning, with potential for improvement (Fig. \ref{fig:results_al_limitations}), as it achieves state-of-the-art results in both domains.

\section*{Acknowledgements}
This work was supported by the Dynamic Data Driven Applications Systems Program, Air Force Office of Scientific Research under Grant FA9550-19-1-0021. We gratefully acknowledge the support of NVIDIA Corporation with the donations of the Titan X and Titan Xp Pascal GPUs used for this research, and DataCrunch.io for the usage of Nvidia V100 GPUs.

\bibliography{dlcitations}

\begin{thebibliography}{46}
\providecommand{\natexlab}[1]{#1}
\providecommand{\url}[1]{\texttt{#1}}
\expandafter\ifx\csname urlstyle\endcsname\relax
  \providecommand{\doi}[1]{doi: #1}\else
  \providecommand{\doi}{doi: \begingroup \urlstyle{rm}\Url}\fi

\bibitem[Alonso et~al.(2021)Alonso, Sabater, Ferstl, Montesano, and
  Murillo]{alonso2021semi}
Inigo Alonso, Alberto Sabater, David Ferstl, Luis Montesano, and Ana~C Murillo.
\newblock Semi-supervised semantic segmentation with pixel-level contrastive
  learning from a class-wise memory bank.
\newblock In \emph{Proceedings of the IEEE/CVF International Conference on
  Computer Vision}, pages 8219--8228, 2021.

\bibitem[Beyer et~al.(2022)Beyer, Zhai, Royer, Markeeva, Anil, and
  Kolesnikov]{beyer2022knowledge}
Lucas Beyer, Xiaohua Zhai, Am{\'e}lie Royer, Larisa Markeeva, Rohan Anil, and
  Alexander Kolesnikov.
\newblock Knowledge distillation: A good teacher is patient and consistent.
\newblock In \emph{Proceedings of the IEEE/CVF Conference on Computer Vision
  and Pattern Recognition}, pages 10925--10934, 2022.

\bibitem[Brostow et~al.(2009)Brostow, Fauqueur, and
  Cipolla]{brostow2009semantic}
Gabriel~J Brostow, Julien Fauqueur, and Roberto Cipolla.
\newblock Semantic object classes in video: A high-definition ground truth
  database.
\newblock \emph{Pattern Recognition Letters}, 30\penalty0 (2):\penalty0 88--97,
  2009.

\bibitem[Casanova et~al.(2020)Casanova, Pinheiro, Rostamzadeh, and
  Pal]{Casanova2020Reinforced}
Arantxa Casanova, Pedro~O. Pinheiro, Negar Rostamzadeh, and Christopher~J. Pal.
\newblock Reinforced active learning for image segmentation.
\newblock In \emph{International Conference on Learning Representations}, 2020.
\newblock URL \url{https://openreview.net/forum?id=SkgC6TNFvr}.

\bibitem[Chen et~al.(2018{\natexlab{a}})Chen, Papandreou, Kokkinos, Murphy, and
  Yuille]{Chen2018DeepLabSI}
Liang-Chieh Chen, George Papandreou, Iasonas Kokkinos, Kevin~P. Murphy, and
  Alan~Loddon Yuille.
\newblock Deeplab: Semantic image segmentation with deep convolutional nets,
  atrous convolution, and fully connected crfs.
\newblock \emph{IEEE Transactions on Pattern Analysis and Machine
  Intelligence}, 40:\penalty0 834--848, 2018{\natexlab{a}}.

\bibitem[Chen et~al.(2018{\natexlab{b}})Chen, Zhu, Papandreou, Schroff, and
  Adam]{Chen_2018_ECCV}
Liang-Chieh Chen, Yukun Zhu, George Papandreou, Florian Schroff, and Hartwig
  Adam.
\newblock Encoder-decoder with atrous separable convolution for semantic image
  segmentation.
\newblock In \emph{Proceedings of the European Conference on Computer Vision
  (ECCV)}, September 2018{\natexlab{b}}.

\bibitem[Chen et~al.(2021)Chen, Yuan, Zeng, and Wang]{Chen_2021_CVPR}
Xiaokang Chen, Yuhui Yuan, Gang Zeng, and Jingdong Wang.
\newblock Semi-supervised semantic segmentation with cross pseudo supervision.
\newblock In \emph{Proceedings of the IEEE/CVF Conference on Computer Vision
  and Pattern Recognition (CVPR)}, pages 2613--2622, June 2021.

\bibitem[Cordts et~al.(2016)Cordts, Omran, Ramos, Rehfeld, Enzweiler, Benenson,
  Franke, Roth, and Schiele]{cordts2016cityscapes}
Marius Cordts, Mohamed Omran, Sebastian Ramos, Timo Rehfeld, Markus Enzweiler,
  Rodrigo Benenson, Uwe Franke, Stefan Roth, and Bernt Schiele.
\newblock The cityscapes dataset for semantic urban scene understanding.
\newblock In \emph{Proceedings of the IEEE conference on computer vision and
  pattern recognition}, pages 3213--3223, 2016.

\bibitem[DeVries and Taylor(2017)]{devries2017improved}
Terrance DeVries and Graham~W Taylor.
\newblock Improved regularization of convolutional neural networks with cutout.
\newblock \emph{arXiv preprint arXiv:1708.04552}, 2017.

\bibitem[Ebrahimi et~al.(2020)Ebrahimi, Gan, Chen, Biamby, Salahi, Laielli,
  Zhu, and Darrell]{ebrahimi2020minimax}
Sayna Ebrahimi, William Gan, Dian Chen, Giscard Biamby, Kamyar Salahi, Michael
  Laielli, Shizhan Zhu, and Trevor Darrell.
\newblock Minimax active learning.
\newblock \emph{arXiv preprint arXiv:2012.10467}, 2020.

\bibitem[Falcon et~al.(2019)]{falcon2019pytorch}
William Falcon et~al.
\newblock Pytorch lightning.
\newblock \emph{GitHub. Note: https://github.
  com/PyTorchLightning/pytorch-lightning}, 3\penalty0 (6), 2019.

\bibitem[French et~al.(2020)French, Laine, Aila, Mackiewicz, and
  Finlayson]{french2020semisupervised}
Geoff French, Samuli Laine, Timo Aila, Michal Mackiewicz, and Graham Finlayson.
\newblock Semi-supervised semantic segmentation needs strong, varied
  perturbations, 2020.

\bibitem[Golestaneh and Kitani(2020)]{equal2020}
S.~Alireza Golestaneh and Kris Kitani.
\newblock Importance of self-consistency in active learning for semantic
  segmentation.
\newblock \emph{BMVC}, 2020.

\bibitem[Guan et~al.(2022)Guan, Huang, Xiao, and Lu]{Guan_2022_CVPR}
Dayan Guan, Jiaxing Huang, Aoran Xiao, and Shijian Lu.
\newblock Unbiased subclass regularization for semi-supervised semantic
  segmentation.
\newblock In \emph{Proceedings of the IEEE/CVF Conference on Computer Vision
  and Pattern Recognition (CVPR)}, pages 9968--9978, June 2022.

\bibitem[Haussmann et~al.(2020)Haussmann, Fenzi, Chitta, Ivanecky, Xu, Roy,
  Mittel, Koumchatzky, Farabet, and Alvarez]{haussmann2020scalable}
Elmar Haussmann, Michele Fenzi, Kashyap Chitta, Jan Ivanecky, Hanson Xu, Donna
  Roy, Akshita Mittel, Nicolas Koumchatzky, Clement Farabet, and Jose~M
  Alvarez.
\newblock Scalable active learning for object detection.
\newblock In \emph{2020 IEEE Intelligent Vehicles Symposium (IV)}, pages
  1430--1435. IEEE, 2020.

\bibitem[He et~al.(2016)He, Zhang, Ren, and Sun]{He_2016_CVPR}
Kaiming He, Xiangyu Zhang, Shaoqing Ren, and Jian Sun.
\newblock Deep residual learning for image recognition.
\newblock In \emph{Proceedings of the IEEE Conference on Computer Vision and
  Pattern Recognition (CVPR)}, June 2016.

\bibitem[He et~al.(2019)He, Zhang, Zhang, Zhang, Xie, and Li]{he2019bag}
Tong He, Zhi Zhang, Hang Zhang, Zhongyue Zhang, Junyuan Xie, and Mu~Li.
\newblock Bag of tricks for image classification with convolutional neural
  networks.
\newblock In \emph{Proceedings of the IEEE/CVF Conference on Computer Vision
  and Pattern Recognition}, pages 558--567, 2019.

\bibitem[Hu et~al.(2021)Hu, Wei, Hu, Ye, Cui, and Wang]{hu2021semi}
Hanzhe Hu, Fangyun Wei, Han Hu, Qiwei Ye, Jinshi Cui, and Liwei Wang.
\newblock Semi-supervised semantic segmentation via adaptive equalization
  learning.
\newblock \emph{Advances in Neural Information Processing Systems}, 34, 2021.

\bibitem[Huang et~al.(2016)Huang, Sun, Liu, Sedra, and
  Weinberger]{huang2016deep}
Gao Huang, Yu~Sun, Zhuang Liu, Daniel Sedra, and Kilian~Q Weinberger.
\newblock Deep networks with stochastic depth.
\newblock In \emph{European conference on computer vision}, pages 646--661.
  Springer, 2016.

\bibitem[Khosla et~al.(2020)Khosla, Teterwak, Wang, Sarna, Tian, Isola,
  Maschinot, Liu, and Krishnan]{khosla2020supervised}
Prannay Khosla, Piotr Teterwak, Chen Wang, Aaron Sarna, Yonglong Tian, Phillip
  Isola, Aaron Maschinot, Ce~Liu, and Dilip Krishnan.
\newblock Supervised contrastive learning.
\newblock \emph{Advances in Neural Information Processing Systems},
  33:\penalty0 18661--18673, 2020.

\bibitem[Li and Alstr{\o}m(2020)]{li2020uncertainty}
Bo~Li and Tommy~Sonne Alstr{\o}m.
\newblock On uncertainty estimation in active learning for image segmentation.
\newblock \emph{arXiv preprint arXiv:2007.06364}, 2020.

\bibitem[Liu et~al.(2021)Liu, Zhi, Johns, and Davison]{liu2021bootstrapping}
Shikun Liu, Shuaifeng Zhi, Edward Johns, and Andrew~J Davison.
\newblock Bootstrapping semantic segmentation with regional contrast.
\newblock \emph{arXiv preprint arXiv:2104.04465}, 2021.

\bibitem[Liu et~al.(2017)Liu, De~Mello, Gu, Zhong, Yang, and
  Kautz]{liu2017learning}
Sifei Liu, Shalini De~Mello, Jinwei Gu, Guangyu Zhong, Ming-Hsuan Yang, and Jan
  Kautz.
\newblock Learning affinity via spatial propagation networks.
\newblock \emph{Advances in Neural Information Processing Systems}, 30, 2017.

\bibitem[Liu et~al.(2022)Liu, Tian, Chen, Liu, Belagiannis, and
  Carneiro]{liu2021perturbed}
Yuyuan Liu, Yu~Tian, Yuanhong Chen, Fengbei Liu, Vasileios Belagiannis, and
  Gustavo Carneiro.
\newblock Perturbed and strict mean teachers for semi-supervised semantic
  segmentation.
\newblock 2022.

\bibitem[Mittal et~al.(2019)Mittal, Tatarchenko, and Brox]{mittal2019semi}
Sudhanshu Mittal, Maxim Tatarchenko, and Thomas Brox.
\newblock Semi-supervised semantic segmentation with high-and low-level
  consistency.
\newblock \emph{IEEE transactions on pattern analysis and machine
  intelligence}, 2019.

\bibitem[Olsson et~al.(2021)Olsson, Tranheden, Pinto, and
  Svensson]{Olsson_2021_WACV}
Viktor Olsson, Wilhelm Tranheden, Juliano Pinto, and Lennart Svensson.
\newblock Classmix: Segmentation-based data augmentation for semi-supervised
  learning.
\newblock In \emph{Proceedings of the IEEE/CVF Winter Conference on
  Applications of Computer Vision (WACV)}, pages 1369--1378, January 2021.

\bibitem[Ouali et~al.(2020)Ouali, Hudelot, and Tami]{Ouali_2020_CVPR}
Yassine Ouali, Celine Hudelot, and Myriam Tami.
\newblock Semi-supervised semantic segmentation with cross-consistency
  training.
\newblock In \emph{Proceedings of the IEEE/CVF Conference on Computer Vision
  and Pattern Recognition (CVPR)}, June 2020.

\bibitem[Paszke et~al.(2019)Paszke, Gross, Massa, Lerer, Bradbury, Chanan,
  Killeen, Lin, Gimelshein, Antiga, Desmaison, Kopf, Yang, DeVito, Raison,
  Tejani, Chilamkurthy, Steiner, Fang, Bai, and Chintala]{paszke2019pytorch}
Adam Paszke, Sam Gross, Francisco Massa, Adam Lerer, James Bradbury, Gregory
  Chanan, Trevor Killeen, Zeming Lin, Natalia Gimelshein, Luca Antiga, Alban
  Desmaison, Andreas Kopf, Edward Yang, Zachary DeVito, Martin Raison, Alykhan
  Tejani, Sasank Chilamkurthy, Benoit Steiner, Lu~Fang, Junjie Bai, and Soumith
  Chintala.
\newblock Pytorch: An imperative style, high-performance deep learning library.
\newblock In H.~Wallach, H.~Larochelle, A.~Beygelzimer, F.~d\textquotesingle
  Alch\'{e}-Buc, E.~Fox, and R.~Garnett, editors, \emph{Advances in Neural
  Information Processing Systems 32}, pages 8024--8035. Curran Associates,
  Inc., 2019.
\newblock URL
  \url{http://papers.neurips.cc/paper/9015-pytorch-an-imperative-style-high-performance-deep-learning-library.pdf}.

\bibitem[Rangnekar et~al.(2022)Rangnekar, Kanan, and
  Hoffman]{rangnekar2022semantic}
Aneesh Rangnekar, Christopher Kanan, and Matthew Hoffman.
\newblock Semantic segmentation with active semi-supervised learning.
\newblock \emph{arXiv preprint arXiv:2203.10730}, 2022.

\bibitem[Sandler et~al.(2018)Sandler, Howard, Zhu, Zhmoginov, and
  Chen]{Sandler_2018_CVPR}
Mark Sandler, Andrew Howard, Menglong Zhu, Andrey Zhmoginov, and Liang-Chieh
  Chen.
\newblock Mobilenetv2: Inverted residuals and linear bottlenecks.
\newblock In \emph{Proceedings of the IEEE Conference on Computer Vision and
  Pattern Recognition (CVPR)}, June 2018.

\bibitem[Shin et~al.(2021)Shin, Xie, and Albanie]{Shin_2021_ICCV}
Gyungin Shin, Weidi Xie, and Samuel Albanie.
\newblock All you need are a few pixels: Semantic segmentation with pixelpick.
\newblock In \emph{Proceedings of the IEEE/CVF International Conference on
  Computer Vision (ICCV) Workshops}, pages 1687--1697, October 2021.

\bibitem[Siddiqui et~al.(2020)Siddiqui, Valentin, and
  Nie{\ss}ner]{siddiqui2020viewal}
Yawar Siddiqui, Julien Valentin, and Matthias Nie{\ss}ner.
\newblock Viewal: Active learning with viewpoint entropy for semantic
  segmentation.
\newblock In \emph{Proceedings of the IEEE/CVF Conference on Computer Vision
  and Pattern Recognition}, pages 9433--9443, 2020.

\bibitem[Sinha et~al.(2019)Sinha, Ebrahimi, and Darrell]{sinha2019variational}
Samarth Sinha, Sayna Ebrahimi, and Trevor Darrell.
\newblock Variational adversarial active learning.
\newblock In \emph{Proceedings of the IEEE/CVF International Conference on
  Computer Vision}, pages 5972--5981, 2019.

\bibitem[Srivastava et~al.(2014)Srivastava, Hinton, Krizhevsky, Sutskever, and
  Salakhutdinov]{srivastava14a}
Nitish Srivastava, Geoffrey Hinton, Alex Krizhevsky, Ilya Sutskever, and Ruslan
  Salakhutdinov.
\newblock Dropout: A simple way to prevent neural networks from overfitting.
\newblock \emph{Journal of Machine Learning Research}, 15\penalty0
  (56):\penalty0 1929--1958, 2014.
\newblock URL \url{http://jmlr.org/papers/v15/srivastava14a.html}.

\bibitem[Tarvainen and Valpola(2017)]{antti2017meanteacher}
Antti Tarvainen and Harri Valpola.
\newblock Mean teachers are better role models: Weight-averaged consistency
  targets improve semi-supervised deep learning results.
\newblock In \emph{Proceedings of the 31st International Conference on Neural
  Information Processing Systems}, NIPS'17, page 1195–1204, Red Hook, NY,
  USA, 2017. Curran Associates Inc.
\newblock ISBN 9781510860964.

\bibitem[Wang et~al.(2021)Wang, Zhou, Yu, Dai, Konukoglu, and
  Van~Gool]{Wang_2021_ICCV}
Wenguan Wang, Tianfei Zhou, Fisher Yu, Jifeng Dai, Ender Konukoglu, and Luc
  Van~Gool.
\newblock Exploring cross-image pixel contrast for semantic segmentation.
\newblock In \emph{Proceedings of the IEEE/CVF International Conference on
  Computer Vision (ICCV)}, pages 7303--7313, 2021.

\bibitem[Wang et~al.(2022)Wang, Wang, Shen, Fei, Li, Jin, Wu, Zhao, and
  Le]{wang2022semi}
Yuchao Wang, Haochen Wang, Yujun Shen, Jingjing Fei, Wei Li, Guoqiang Jin,
  Liwei Wu, Rui Zhao, and Xinyi Le.
\newblock Semi-supervised semantic segmentation using unreliable pseudo labels.
\newblock In \emph{Proceedings of the IEEE/CVF International Conference on
  Computer Vision and Pattern Recognition (CVPR)}, 2022.

\bibitem[Xie et~al.(2020{\natexlab{a}})Xie, Luong, Hovy, and Le]{xie2020self}
Qizhe Xie, Minh-Thang Luong, Eduard Hovy, and Quoc~V Le.
\newblock Self-training with noisy student improves imagenet classification.
\newblock In \emph{Proceedings of the IEEE/CVF conference on computer vision
  and pattern recognition}, pages 10687--10698, 2020{\natexlab{a}}.

\bibitem[Xie et~al.(2020{\natexlab{b}})Xie, Feng, Chen, Sun, Ma, and
  Song]{xie2020deal}
Shuai Xie, Zunlei Feng, Ying Chen, Songtao Sun, Chao Ma, and Mingli Song.
\newblock Deal: Difficulty-aware active learning for semantic segmentation.
\newblock In \emph{Proceedings of the Asian Conference on Computer Vision},
  2020{\natexlab{b}}.

\bibitem[Yang et~al.(2022)Yang, Zhuo, Qi, Shi, and Gao]{yang2021st++}
Lihe Yang, Wei Zhuo, Lei Qi, Yinghuan Shi, and Yang Gao.
\newblock St++: Make self-training work better for semi-supervised semantic
  segmentation.
\newblock In \emph{Proceedings of the IEEE/CVF International Conference on
  Computer Vision and Pattern Recognition (CVPR)}, 2022.

\bibitem[Yun et~al.(2019)Yun, Han, Oh, Chun, Choe, and Yoo]{yun2019cutmix}
Sangdoo Yun, Dongyoon Han, Seong~Joon Oh, Sanghyuk Chun, Junsuk Choe, and
  Youngjoon Yoo.
\newblock Cutmix: Regularization strategy to train strong classifiers with
  localizable features.
\newblock In \emph{Proceedings of the IEEE/CVF International Conference on
  Computer Vision}, pages 6023--6032, 2019.

\bibitem[Zhang et~al.(2021)Zhang, Torr, Ranftl, and Richter]{zhang2021looking}
Feihu Zhang, Philip Torr, Ren{\'e} Ranftl, and Stephan Richter.
\newblock Looking beyond single images for contrastive semantic segmentation
  learning.
\newblock \emph{Advances in neural information processing systems},
  34:\penalty0 3285--3297, 2021.

\bibitem[Zhao et~al.(2019)Zhao, Wang, Yang, and Cai]{zhao2019region}
Shuai Zhao, Yang Wang, Zheng Yang, and Deng Cai.
\newblock Region mutual information loss for semantic segmentation.
\newblock \emph{Advances in Neural Information Processing Systems}, 32, 2019.

\bibitem[Zhou et~al.(2022)Zhou, Wang, Konukoglu, and
  Van~Gool]{zhou2022rethinking}
Tianfei Zhou, Wenguan Wang, Ender Konukoglu, and Luc Van~Gool.
\newblock Rethinking semantic segmentation: A prototype view.
\newblock In \emph{Proceedings of the IEEE/CVF Conference on Computer Vision
  and Pattern Recognition}, pages 2582--2593, 2022.

\bibitem[Zhou et~al.(2021)Zhou, Xu, Zhang, Gao, and Heng]{zhou2021c3}
Yanning Zhou, Hang Xu, Wei Zhang, Bin Gao, and Pheng-Ann Heng.
\newblock C3-semiseg: Contrastive semi-supervised segmentation via cross-set
  learning and dynamic class-balancing.
\newblock In \emph{Proceedings of the IEEE/CVF International Conference on
  Computer Vision}, pages 7036--7045, 2021.

\bibitem[Zhu et~al.(2020)Zhu, Zhang, Wu, Zhang, He, Zhang, Manmatha, Li, and
  Smola]{zhu2020improving}
Yi~Zhu, Zhongyue Zhang, Chongruo Wu, Zhi Zhang, Tong He, Hang Zhang,
  R~Manmatha, Mu~Li, and Alexander Smola.
\newblock Improving semantic segmentation via self-training.
\newblock \emph{arXiv preprint arXiv:2004.14960}, 2020.

\end{thebibliography}

\clearpage
\begin{center}
    {\Large Supplemental Material \normalsize}
\end{center}
\beginsupplement

We will release all code towards reproducing the results in this paper post publication. We provide a full summary of our experiments on semi-supervised learning in Table \ref{tab:trainingschedule_ssl}. Table \ref{tab:semisupervisedlearning_cityscapes} reports the results on CityScapes for semi-supervised learning as a comparative analysis to previous state-of-the-art and class-wise performance on the IoU metric. Tables \ref{tab:activelearning_classwise_camvid} and \ref{tab:activelearning_classwise_cityscapes} report the class-wise IoU performance on CamVid and CityScapes datasets for Active learning respectively. We train all our models on 4 Nvidia V100 GPUs with 16GB, and use Pytorch \cite{paszke2019pytorch} with Pytorch Lightning \cite{falcon2019pytorch} for all our experiments. We also use 16-bit precision as needed, especially for training ResNet101-DeepStem encoder based DeepLabv3+ models.

\begin{table}[]
\centering
\def\arraystretch{1.2}
\caption{Summary of all Semi-supervised learning experiments on the CityScapes datasets. LD indicates the total number of labeled data images used following previous state-of-the-art protocols. We use a 1:1.5 labeled to unlabeled image ratio as that is the maximum we are able to fit on the GPU, and use contrastive representation learning for all experiments. We report the results on the mIoU metric.}
\label{tab:trainingschedule_ssl}
\resizebox{\columnwidth}{!}{%
\begin{tabular}{llllllll}
Dataset & Experiment & Epochs & \begin{tabular}[c]{@{}l@{}}Batch\\ Size\\ (Labeled)\end{tabular} & \begin{tabular}[c]{@{}l@{}}Batch\\ Size\\ (Unlabeled)\end{tabular} & \begin{tabular}[c]{@{}l@{}}Iterations\\ Per Epoch\end{tabular} & \begin{tabular}[c]{@{}l@{}}mIoU\\ Run 1\end{tabular} & \begin{tabular}[c]{@{}l@{}}mIoU\\ Run 2\end{tabular} \\ \midrule
CityScapes & R50, 100 LD & 240 & 2 & 3 & 400 & 59.99 & 61.00 \\
 & R50, 372 LD & 240 & 2 & 3 & 200 & 72.92 & 73.56 \\
 & R50, 744 LD & 240 & 2 & 3 & 200 & 74.75 & 75.34 \\ \midrule
CityScapes & R101D, 186 LD & 240 & 2 & 3 & 400 & 70.89 & 71.89 \\
 & R101D, 372 LD & 240 & 2 & 3 & 300 & 76.21 & 76.88 \\
 & R101D, 744 LD & 240 & 2 & 3 & 300 & 77.86 & 78.39 \\
 & R101D, 1488 LD & 240 & 2 & 3 & 400 & 79.58 & 79.64 \\ \bottomrule
\end{tabular}
}
\end{table}

\begin{table}[t]
\caption{Semi-supervised learning results on the CityScapes dataset in terms of labeled images (the first column indicates the ratio of data usage, and the second column indicates the number of labeled samples). All numbers are reported on mIoU metric using a ResNet-101-DeepStem as the encoder with DeepLab-v3+, except those marked with *, which indicates a ResNet-50 encoder to ensure fairness \cite{yang2021st++}. SupOnly stands for supervised training on the portion of labeled data, and SupOnly + Rep indicates a representation learning head is added to the decoder along with the prediction head. \label{tab:semisupervisedlearning_cityscapes}}
\tiny
\def\arraystretch{1.2}
\begin{center}
\resizebox{\columnwidth}{!}{%
\begin{tabular}{@{}lllllllllll@{}}
 & & SupOnly & \begin{tabular}[c]{@{}l@{}}SupOnly\\ + Rep\end{tabular} & ST++ \cite{yang2021st++} & \begin{tabular}[c]{@{}l@{}}Mean\\ Teacher \cite{antti2017meanteacher}\end{tabular} & CCT \cite{Ouali_2020_CVPR} & CPS \cite{Chen_2021_CVPR} & AEL \cite{hu2021semi} & U$^2$PL \cite{wang2022semi} & \textbf{S4AL+} \\ \midrule
1/30 & 100 & 56.25* & 58.29* & 61.4* & - & - & - & - & - & 61.00* \\
1/16 & 186 & 66.1 & 67.16 & - & 69.03 & 69.32 & 69.72 & 74.45 & 74.90 & 71.89 \\
1/8 & 372 & 66.5* & 68.3* & 72.7* & - & - & - & - & - & 73.56* \\
 & & 71.08 & 72.22 & - & 72.06 & 74.12 & 74.31 & 75.55 & 76.48 & 76.88 \\
1/4 & 744 & 68.7* & 70.98* & 73.8* & - & - & - & - & - & 75.34* \\
 & & 72.53 & 73.78 & - & 74.20 & 75.99 & 74.58 & 77.48 & 78.51 & 78.39 \\
1/2 & 1488 & 76.75 & 77.42 & - & 78.15 & 78.10 & 76.81 & 79.01 & 79.12 & 79.64 \\ \bottomrule
\end{tabular}%
}
\end{center}
\end{table}

\begin{table}[t]
\caption{Active learning results on the CamVid dataset compared to the existing state of art, highlighting the total amount of data required. The results are reported on the IoU metric, and the numbers within the brackets () indicate the total labeled data usage.}
\label{tab:activelearning_classwise_camvid}
\small
\def\arraystretch{1.2}
\begin{center}
\resizebox{\columnwidth}{!}{%
\begin{tabular}{@{}lllllllllllll@{}}
 & Sky & Building & \begin{tabular}[c]{@{}l@{}}Column\\ Pole\end{tabular} & Road & \begin{tabular}[c]{@{}l@{}}Side\\ walk\end{tabular} & Tree & Sign & Fence & Car & Pedestrian & Bicyclist & mIoU \\ \midrule
S4AL \cite{rangnekar2022semantic} (13.8\%) & 91.51 & 82.30 & 15.54 & 91.52 & 72.05 & 73.59 & 40.11 & 31.17 & 78.99 & 48.65 & 49.55 & 61.36 \\
\textbf{S4AL+} (12.1\%) & 91.08 & 81.99 & 16.62 & 93.88 & 79.21 & 75.65 & 37.39 & 34.27 & 80.31 & 51.20 & 47.56 & 62.65 \\
Supervised (100.0\%) & 91.48 & 81.36 & 20.84 & 93.90 & 79.30 & 75.17 & 40.58 & 33.76 & 79.97 & 49.83 & 52.86 & 63.55 \\ \bottomrule
\end{tabular}%
}
\end{center}
\end{table}

\begin{table}[t]
\caption{Active learning results on the CityScapes dataset compared to the existing state of art, highlighting the total amount of data required. The results are reported on the IoU metric, and the numbers within the brackets () indicate the total labeled data usage. \label{tab:activelearning_classwise_cityscapes}}
\small
\def\arraystretch{1.2}
\begin{center}
\resizebox{\columnwidth}{!}{%
\begin{tabular}{@{}lllllllllll@{}}
 & Road & \begin{tabular}[c]{@{}l@{}}Side\\ walk\end{tabular} & Building & Wall & Fence & Pole & \begin{tabular}[c]{@{}l@{}}Traffic\\ Light\end{tabular} & \begin{tabular}[c]{@{}l@{}}Traffic\\ Sign\end{tabular} & Vegetation & Terrain \\ \midrule
DEAL \cite{xie2020deal} (40\%) & 95.89 & 71.69 & 87.09 & 45.61 & 44.94 & 38.29 & 36.51 & 55.47 & 87.53 & 56.90 \\
S4AL \cite{rangnekar2022semantic} (16.3\%) & 96.53 & 80.16 & 86.63 & 48.15 & 46.41 & 35.10 & 43.78 & 58.07 & 88.91 & 61.50 \\
\textbf{S4AL+} (15.1\%) & 97.38 & 79.07 & 87.87 & 43.69 & 46.47 & 32.24 & 41.94 & 57.35 & 89.14 & 57.89 \\
Supervised (100\%) & 97.58 & 80.55 & 88.43 & 51.22 & 47.61 & 35.19 & 42.19 & 56.79 & 89.41 & 60.22 \\ \\
 & Sky & Pedestrian & Rider & Car & Truck & Bus & Train & Motorcycle & Bicycle & mIoU \\ \midrule
DEAL \cite{xie2020deal} (40\%) & 91.78 & 64.25 & 39.77 & 88.11 & 56.87 & 64.46 & 50.39 & 38.92 & 56.59 & 61.64 \\
S4AL \cite{rangnekar2022semantic} (16.3\%) & 92.08 & 65.14 & 39.75 & 90.52 & 64.15 & 65.31 & 41.53 & 46.18 & 58.97 & 63.62 \\
\textbf{S4AL+} (15.1\%) & 92.47 & 67.23 & 43.44 & 90.34 & 49.47 & 66.37 & 54.72 & 47.34 & 64.01 & 63.60 \\ 
Supervised (100\%) & 92.69 & 65.12 & 37.32 & 90.67 & 66.24 & 71.84 & 63.84 &42.35 & 61.84 & 65.30 \\ \bottomrule
\end{tabular}%
}
\end{center}
\end{table}

\end{document}